%% file: paper.tex
\documentclass{article}
\usepackage{spconf,amsmath,graphicx}
\usepackage[numbers]{natbib}
\usepackage{tikz}
\usepackage{pgfplots}
\usepackage{amsmath}
\usepackage{amssymb}

\newcommand{\vect}[1]{\boldsymbol{#1}}
\usetikzlibrary{pgfplots.groupplots}
\usetikzlibrary{matrix}

\title{Pruning Convolutional Neural Networks for Image Instance Retrieval}

\name{
\begin{tabular}{c}
Gaurav Manek\sthanks{The first three authors contributed equally.}$^{1}$, Jie Lin\footnotemark[1]$^{1}$, Vijay Chandrasekhar\footnotemark[1]$^{1}$, \\
Lingyu Duan$^{2}$, Sateesh Giduthuri$^{1}$, Xiaoli Li$^{1}$, Tomaso Poggio$^{3}$
\end{tabular}
}

\address{Institute for Infocomm Research, Singapore$^{1}$ \quad Peking University, China$^{2}$ \quad MIT, USA$^{3}$}

\begin{document}
\ninept
\maketitle
\begin{abstract}
In this work, we focus on the problem of image instance retrieval with deep descriptors extracted from \textbf{pruned} Convolutional Neural Networks (CNN).
The objective is to heavily prune convolutional edges while maintaining retrieval performance.
To this end, we introduce both data-independent and data-dependent heuristics to prune convolutional edges,
and evaluate their performance across various compression rates with different deep descriptors over several benchmark datasets.
Further, we present an end-to-end framework to fine-tune the pruned network, with a triplet loss function specially designed for the retrieval task.
We show that the combination of heuristic pruning and fine-tuning offers $5\times$ compression rate without considerable loss in retrieval performance.
\end{abstract}

\section{Introduction}
\label{sec:intro}

Image instance retrieval is the problem of retrieving images from a large database that contain or depict similar objects to a target image. 
Convolutional Neural Networks (CNN)-based descriptors \cite{babenko2014neural,babenko2,crow,ToliasSJ15,wan2014deep} have recently been used to generate compact image descriptors with high retrieval performance, 
and are rapidly becoming the dominant approach for retrieval problem. Their major drawback is the size of their models, running into hundreds of megabytes.

Smaller networks are desirable for use in mobile and embedded applications, where storage, transmission, and computational power is limited. 
For efficient hardware implementations of deep neural networks, smaller networks reduce cost and improve chip performance. 
Storing the entire network on-chip allows fast access and reduce processing latency. 
There are also gains in distributed training, where network latency bottlenecks the sharing of updated parameters, and smaller networks significantly improve speed. 
Finally, emerging MPEG standards like Compact Descriptors for Visual Search (CDVS) \cite{duan2016overview} and 
Compact Descriptors for Video Analysis (CDVA) \cite{chandrasekhar2016compression} require memory-efficient models for streaming and hardware implementations.

Model compression strategies reduce the computational, memory, and bandwidth costs; and pruning is a common first technique. 
Pruning algorithms reduce network size by discarding edges or nodes, and can be heuristic or analytic \cite{reed1993pruning}. 
Modern pruning algorithms have thus far been evaluated on CNN performing image classification with a softmax loss function; 
we investigate the adaptation of these algorithms to image instance retrieval problem, and present an empirical evaluation of their effectiveness.

\subsection{Related Work}
\label{sec:literature}

\textbf{Image Instance Retrieval with CNN}.
Image retrieval systems generally construct a \emph{global image descriptor}, a vector that represents the contents of an image.
Instead of descriptor extracted from fully-connected layer, 
state-of-the-art use the intermediate output of CNN (e.g. convolutional layers) with additional pooling operations to generate descriptors \cite{babenko2014neural,babenko2,crow}.
The application of multi-scale and multi-rotation feature construction and pooling further improves scale- and rotation-invariance \cite{ToliasSJ15, MorereVLPCP16}.
Very recently, pre-trained CNNs for ImageNet classification are repurposed for the image retrieval problem,
by fine-tuning them with retrieval specific loss functions such as triplet loss~\cite{tinydcc,Arandjelovic16,Gordo1}.

\textbf{Network Pruning}
Heuristic pruning algorithms generally assign either nodes or edges \emph{salience} scores and remove those with the lowest scores \cite{reed1993pruning}. 
In early neural network literature there was a considerable interest in developing such algorithms. 
Heuristics include \emph{Optimal Brain Damage} by \citet{lecun1989optimal}; derivative-based methods by \citet{mozer1989skeletonization}, and \citet{karnin1990simple}. 
Other work removes nodes where the weights of incoming connections has the smallest variance, but this has been mostly studied in the realm of fault tolerance \citep{segee1991fault}.

In current networks, a popular strategy is the removal of low-magnitude edges. \citet{han2015learning} demonstrates this simple strategy on \texttt{AlexNet}, 
reducing the number of free parameters by a factor of $9\times$ by pruning alone without any loss in image classification performance.
Their reported results heavily prune edges from fully-connected layers ($\sim$89\% are pruned); on convolutional layers alone, 37\% of convolutional edges are dropped.
Thus, the majority of the savings comes from the fully-connected layers, which contain 96\% of the parameters of the entire network.

There are also analytic algorithms compress networks layer-by-layer, replacing the convolution matrix with a compressed representation. 
This compression is achieved by removing redundancy in the function that each layer computes. 
This approach has been vigorously explored in recent years, with work by \citet{kim2015compression}, \citet{mariet2016diversity}, \citet{vadim2015speeding}, 
and \citet{denton2014exploiting} each with different redundancy reduction mechanisms.

\subsection{Contributions}
Compared to fully-connected layers that contain the most redundancy, heavily pruning edges on convolutional layers is more challenging.
In this work, we focus on heuristic criteria to prune convolutional edges especially for image instance retrieval.
We make the following contributions:
\begin{itemize}
\item We investigate both data-independent and data-dependent heuristics to prune convolutional edges.
We perform a thorough evaluation across various compression rates and deep pooled descriptors over several benchmark datasets.
Results suggest that heuristic pruning is capable of reducing the network size by $2\times$ without retrieval performance loss.
 
\item We introduce an end-to-end framework for fine-tuning pruned network, specially tailored to image instance retrieval with a triplet loss function.
Combining pruning and rank-based fine-tuning can provide a factor of $5\times$ compression with minimal loss in retrieval performance.
\end{itemize}

\section{Method}
\label{sec:methodology}
\subsection{Pruning Convolutional Edges}
We consider different heuristics to assign salience scores\cite{reed1993pruning} to edges.
Consider an arbitrary layer in a neural network,
$N_i$ is a random variable following the activation of the $i^\text{th}$ node when presented with data from a training set,
$w_{i,j}$ is the weight of the edge connecting that node to the $j^\text{th}$ node in the next layer,
$\mathcal{L}$ is the value of the loss function when run on some data.
From existing literature, we consider computing the following heuristics for each edge:
\begin{enumerate}
\item $|w_{i,j}|$, the simplest heuristic, requiring no data. This was recently popularized by \citet{han2015learning}.

\item $\frac{\text{d} \mathcal{L}}{\text{d} w_{i,j}} w_{i,j}$, which was recently used by \citet{MolchanovTKAK16} to prune networks for transfer learning. 
This is justified as the Taylor expansion of the function computing the difference in loss function with and without the weight $w_{i,j}$, and is shown to be applicable to transfer learning.

\item $\langle |N_i| \rangle \cdot |w_{i,j}|$ Mean activation.

\item $\text{Var}[N_i] \cdot w_{i,j}^2$ Variance of activation.
\end{enumerate}
One may notice that heuristic 2 requires data with ground-truth labels to compute the loss-function term ($\frac{\text{d} \mathcal{L}}{\text{d} w_{i,j}}$). 
Computing heuristic 3 and 4 requires data, but does not require that it have labels.

In order to prune the network to a fraction $t$ of its original size, we first compute the salience score for each edge in the network. 
We then sort all salience scores across all layers and select the threshold salience value, $\tau$, such that $(1 - t)$ of the salience scores are below this value. 
We then remove all edges with salience scores less than $\tau$.
Throughout our experiments, we report the network size as the total fraction of edges removed. We do not prune bias nodes or report them in the network size.

\subsection{Convolutional Feature Pooling}
In constructing the global image descriptor we append a pooling layer to the pruned network.
The pooling function employed is critical to the performance of the model.
In this work, we consider Square-root pooling (SQP)~\cite{linjie2017} and Regional-Maximum Activations of Convolutions (R-MAC)~\cite{ToliasSJ15} pooling functions.
Consider an arbitrary image $X$, with $C$ feature maps $\{\vect{x}_1,...,\vect{x}_C\}$ extracted from intermediate layer, $\vect{x}_c$ is a feature map of width $W$ and height $H$.
Square-root pooling, $f_{2}(\cdot)$ is defined as
\begin{equation}
f^\text{SQP}(\vect{x}_c) = \sqrt{\frac{1}{W\cdot H} \sum^{W\cdot H}_{i=1} x_{c,i}^{2}}.
\label{eq:pool}
\end{equation}

R-MAC~\cite{ToliasSJ15} pooling is computed by first performing maximum pooling over regions of interest (ROI), then average pooling.
\begin{equation}
f^\text{R-MAC}(\vect{x}_c) = \frac{1}{N_\text{ROI}} \sum^{N_\text{ROI}}_{i=1} \max_{j \in [1,S_\text{ROI}]} (x^{i}_{c,j}),
\label{eq:roipool}
\end{equation}
where $S_\text{ROI}$ is the number of ROIs, and $\vect{x}^{i}_{c}$ denotes the $i^{th}$ ROI sampled from feature map $\vect{x}_{c}$, with size $S_\text{ROI} \leq W\cdot H$.

\subsection{Triplet-based Fine-tuning}
To fine-tune remaining parameters in an end-to-end manner, we need to design a loss function for the pruned network.
Following existing work by \citet{Arandjelovic16,Gordo1}, we define a triplet $(X^q,X^{+},X^{-})$ that contains the query image $X^q$, a positive matching image $X^{+}$ and a negative, non-matching image $X^{-}$. 
The images are selected so that query image $X^q$ is more similar to positive image $X^{+}$ than to the negative image $X^{-}$. 
The triplet should meet the condition that $K(X^q, X^{+}) > K(X^q, X^{-})$, where $K$ is a function computing pairwise image similarity.

Accordingly, we define the triplet loss as:
\begin{equation}
L_{q,+,-}= \max\{0, m + K(X^q, X^{-}) - K(X^q, X^{+})\},
\label{eq:triplet_loss}
\end{equation}
where $m$ is a positive margin parameter.

Following~\cite{babenko2}, we define the similarity measure $K$ as:
\begin{equation}
K(X,Y) = \beta(X)\beta(Y) \sum^C_{c=1} k(f(\vect{x}_c),f(\vect{y}_c)),
\label{eq:match}
\end{equation}
where $f(.)$ denotes the pooling operation applied on feature maps, 
$k(f(\vect{x}_c),f(\vect{y}_c)) = <f(\vect{x}_c),f(\vect{y}_c)>$ is the scalar product of the pooled features, 
$\beta(.)$ is a normalization term computed by $\beta(X)=\sqrt{\sum^C_{c=1} k(f(\vect{x}_c),f(\vect{x}_c))}$.

\section{Experiments}
\label{sec:experiments}
We begin with a \texttt{VGG-VeryDeep-16} network \cite{SimonyanZ14a} pre-trained on ImageNet, and keep only the layers from the input up to and including the last pooling layer \texttt{pool5}. 
Convolutional layers are pruned using each of the four heuristics, then fine-tuned for 20 epochs on the image retrieval task using the triplet loss function discussed earlier.
All pruning and fine-tuning, are implemented with the MatConvNet library, with the \texttt{3D-Landmarks} \cite{Radenovic-ECCV16} dataset. 
Testing is performed on the \texttt{INRIA Holidays} \cite{jegou2008hamming}, \texttt{Oxford5k} \cite{Philbin07} and \texttt{Paris6k} \cite{Philbin08} datasets which consist of outdoor scenes and buildings; 
and the \texttt{UKbench} \cite{nister-stewenius-cvpr-2006} dataset which features close-up shots of objects in indoor environments.

In all reported results, the accuracy metric for the \texttt{Holidays}, \texttt{Oxford5k}, and \texttt{Paris6k} datasets is the mean average precision (MAP), 
and the metric for the \texttt{UKbench} dataset is $4\times$recall@4. 
Note that we report results without post-processing (e.g., PCA whitening) on pooled features.

\input{results_overview_noft.tex}

\subsection{Pooling Features}
We first pruned the convolutional layers of \texttt{VGG-VeryDeep-16} using the four heuristics discussed earlier. 
Each network was pruned to five different sizes, from 10\% to 50\% of the original network size. 
The performance of each pruned network on each of the four datasets is presented in Figure~\ref{fig:res_overview_noft}.

From Figure~\ref{fig:res_overview_noft}, we observe that heuristic 1 (magnitude of edge) consistently performs better than the other heuristics. 
In fact, until about 40\% of the edges are remaining, networks pruned with heuristic 1 perform not significantly worse than unpruned networks. 
This corresponds to a $2.5\times$ savings in size for minimal computational and implementation effort.

Additionally, we observe that the accuracy graphs of networks seldom intersect, and so heuristic 1 dominates the entire domain. 
This, along with the advantage of requiring minimal computation and no data, suggests that heuristic 1 is better suited to practical implementations than the other heuristics proposed.

We also use the data in Figure~\ref{fig:res_overview_noft} to compare the performance of SQP and R-MAC features. 
Across three out of four datasets, SQP features perform better than networks with R-MAC features pruned to the same size. 
From this, we observe that SQP features are generally superior, providing around 2 percentage points' performance gain over R-MAC.
Thus, we choose SQP for fine-tuning in the subsequent sections.

\input{results_finetuning.tex}

\subsection{Fine-tuning}
As established by \citet{han2015learning,han2015deep}, fine-tuning pruned networks can recover image classification performance lost in pruning. 
To investigate this effect for image instance retrieval, we fine-tuned networks pruned with each heuristic, and then evaluated their performance. 
These networks were pruned to 50\% of their original size, and evaluated with SQP. 
Figure~\ref{fig:finetuning} shows the performance of these networks before and after fine-tuning.

We observe that in two of the four datasets, fine-tuning improves performance significantly, regardless of the pruning heuristic. 
Even after $50\%$ pruning, the mean average precision of these networks approaches $80\%$. 
Also, after fine-tuning, heuristic 1 reports higher performance in all datasets. 
This further supports our earlier recommendation of heuristic 1, and corroborates the work of \citet{han2015learning,han2015deep}.

We note that fine-tuning decreases the $4\times$recall@4 score on the \texttt{UKbench} dataset, regardless of heuristic used. 
This is likely because fine-tuning on the building-centric \texttt{3D-Landmarks} dataset transfers well to \texttt{Oxford5k} and \texttt{Paris6k}, but poorly to the object-centric \texttt{UKbench} dataset.

\input{results_sh_threshold.tex}

\subsection{Pruned Network Size}
We now investigate the size of the network pruned and its effect on accuracy. 
As per our earlier recommendation, we choose several networks pruned (to different sizes) with heuristic 1, pooled with SQP, and fine-tuned. 
We evaluate their performance on each of testing datasets before and after fine-tuning, and present the results in Figure~\ref{fig:res_sh_threshold}.

We note that networks pruned to $20\%$ of the initial model size exhibit poor performance compared to networks pruned to $50\%$, but this gap in performance diminishes upon fine-tuning. 
Fine-tuning allows us to improve compression rate from a factor of $2\times$ to $5\times$ with minimal performance penalty.

This performance improvement does not extend to networks pruned to 10\% of their initial size. 
While these networks generally exhibit improvement in performance, their performance does not reach the performance of networks pruned to $50\%$ of their size. 
This discontinuity suggests that the pruning algorithms are no longer able to exploit redundancy in the model, and further pruning will worsen performance. 
The minimum model size without loss of performance lies between $10-20\%$ of the original model size.

Further investigation in this threshold is warranted. 
We observe this transition at around the same range in all datasets, suggesting that this is a property of the network or the heuristic. 
Further experiments may show that the heuristics perform differently at extremely small sizes and with little redundancy.

\input{layersize.tex}

\subsection{Pruned Layer Sizes}
We chart the size of each convolutional layer in the network in Figure~\ref{fig:layersize} for three different sizes when pruned with heuristic 1. 
As the network size shrinks, layers further up the network lose convolutional edges disproportionately more quickly. 
Even when the network is pruned to 10\% of its original size, the lowest layer still retains 95\% of its edges, and the highest layers retain only about 8\% of their edges. 
This same trend is observed across all heuristics.

\section{Conclusions}
Pruning edges on convolutional layers is a more challenging operation than on fully-connected layers.
In this work, we presented an end-to-end framework for compressing CNN, specially tailored to efficiently pruning convolutional edges with a triplet loss function. 
We present thorough evaluation across varied pruning parameters and deep features on several datasets. 
Our experiments suggest that pruning and fine-tuning can provide a factor of $2\times$ to $5\times$ compression with minimal loss in performance.

\section{References}
\label{sec:ref}

\bibliographystyle{IEEEtranN}
\renewcommand{\section}[2]{}%
\bibliography{refs}

%\vfill
%\pagebreak

\end{document}

%% file: results_overview_noft.tex
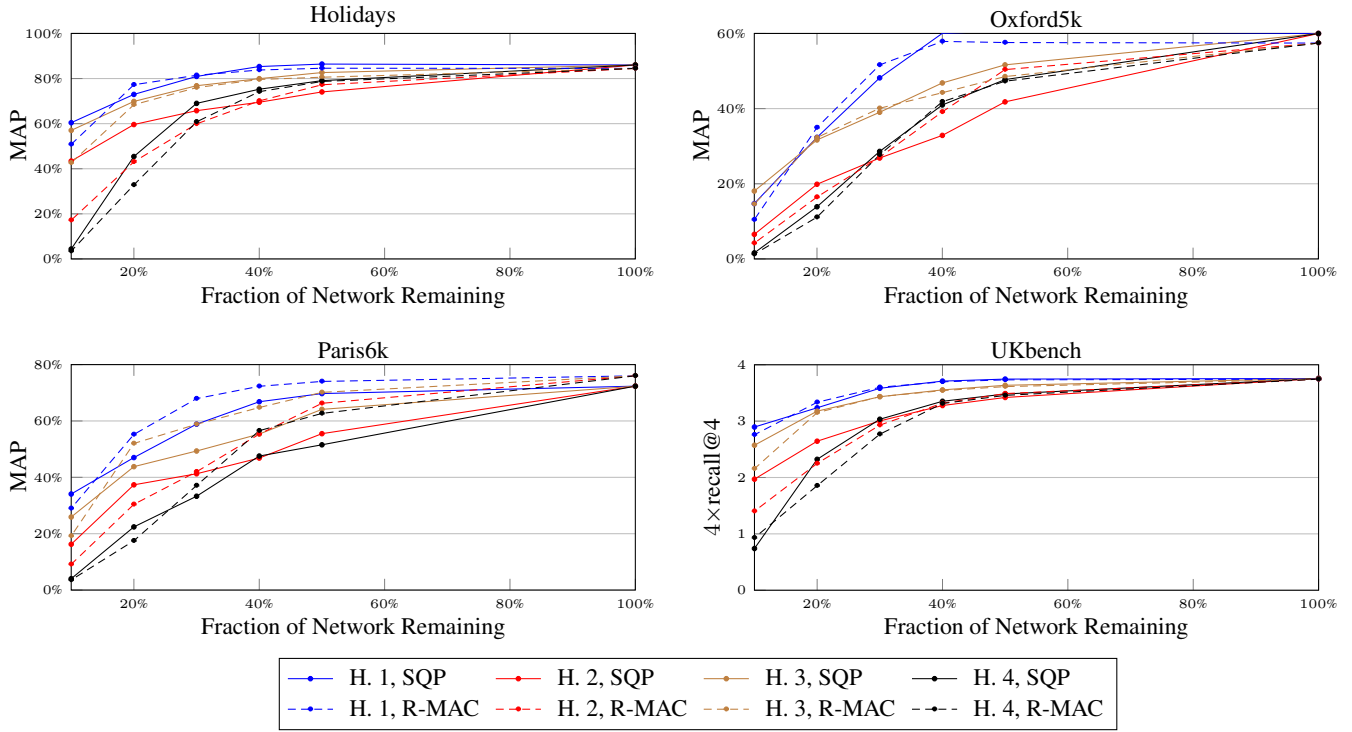
\begin{figure*}[tb]
\begin{minipage}[b]{1.0\linewidth}
  \centering
  \begin{tikzpicture}{

\pgfplotscreateplotcyclelist{custom_color_list}{%
blue,mark=*,mark size=0.85\\%
red,mark=*,mark size=0.85\\%
brown,mark=*,mark size=0.85\\%
black,mark=*,mark size=0.85\\%
blue,densely dashed,mark=*,mark size=0.85\\%
red,densely dashed,mark=*,mark size=0.85\\%
brown,densely dashed,mark=*,mark size=0.85\\%
black,densely dashed,mark=*,mark size=0.85\\%
}

\begin{groupplot}[
    group style={
        group name=results overview,
        group size=2 by 2,
        vertical sep=40pt,
        horizontal sep=45pt,
    },
    title={Title},
    title style={at={(axis description cs:0.5,.92)},anchor=south},
    width = 7.5cm, height = 3cm,
    scale only axis,
    xticklabel={\pgfmathparse{\tick*100}\pgfmathprintnumber{\pgfmathresult}\%},
    xmin=0.1, xmax=1.0,
    x tick label style = {font=\tiny,anchor=north},
    /pgf/bar width=7pt,
    ymajorgrids,
    x label style={at={(axis description cs:0.5,0.09)},anchor=north,font=\small},
    xlabel={Fraction of Network Remaining},
    y label style={at={(axis description cs:0.07,0.7)},anchor=east,font=\small},
    ylabel={MAP},
    point meta={y*100},
    %nodes near coords={\pgfmathprintnumber[fixed,precision=0]\pgfplotspointmeta\%},
    %every node near coord/.append style={font=\scriptsize},
    y tick label style={font=\tiny},
    yticklabel={\pgfmathparse{\tick*100}\pgfmathprintnumber{\pgfmathresult}\%},
    ytick pos=left,
    ymin=0, ymax=1,
    cycle list name=custom_color_list,
    legend cell align=left,
    legend columns=4,
    legend style={
        at={(-0.1,-0.3)},
        anchor=north,
        column sep=1ex
    }
  ]
\nextgroupplot[title={Holidays}]							
\addplot+[] coordinates {	(1, 0.8602)	(0.5, 0.8642)	(0.4, 0.8532)	(0.3, 0.8094)	(0.2, 0.7296)	(0.1, 0.6042)	};
\addplot+[] coordinates {	(1, 0.8602)	(0.5, 0.7406)	(0.4, 0.6949)	(0.3, 0.6574)	(0.2, 0.596)	(0.1, 0.4341)	};
\addplot+[] coordinates {	(1, 0.8602)	(0.5, 0.8265)	(0.4, 0.7995)	(0.3, 0.7681)	(0.2, 0.6985)	(0.1, 0.5701)	};
\addplot+[] coordinates {	(1, 0.8602)	(0.5, 0.7907)	(0.4, 0.7526)	(0.3, 0.6899)	(0.2, 0.4542)	(0.1, 0.0439)	};
\addplot+[] coordinates {	(1, 0.845)	(0.5, 0.8457)	(0.4, 0.8379)	(0.3, 0.8144)	(0.2, 0.7725)	(0.1, 0.5082)	};
\addplot+[] coordinates {	(1, 0.845)	(0.5, 0.7724)	(0.4, 0.7007)	(0.3, 0.5989)	(0.2, 0.4307)	(0.1, 0.1719)	};
\addplot+[] coordinates {	(1, 0.845)	(0.5, 0.8059)	(0.4, 0.7958)	(0.3, 0.7598)	(0.2, 0.6839)	(0.1, 0.4262)	};
\addplot+[] coordinates {	(1, 0.845)	(0.5, 0.787)	(0.4, 0.7418)	(0.3, 0.6081)	(0.2, 0.3282)	(0.1, 0.0347)	};
							
\nextgroupplot[title={Oxford5k},
    ymax=0.6]							
\addplot+[] coordinates {	(1, 0.5998)	(0.5, 0.6063)	(0.4, 0.6003)	(0.3, 0.4818)	(0.2, 0.3235)	(0.1, 0.1473)	};
\addplot+[] coordinates {	(1, 0.5998)	(0.5, 0.4179)	(0.4, 0.3287)	(0.3, 0.2685)	(0.2, 0.1987)	(0.1, 0.0654)	};
\addplot+[] coordinates {	(1, 0.5998)	(0.5, 0.5164)	(0.4, 0.4683)	(0.3, 0.3903)	(0.2, 0.3166)	(0.1, 0.1806)	};
\addplot+[] coordinates {	(1, 0.5998)	(0.5, 0.4772)	(0.4, 0.4097)	(0.3, 0.2861)	(0.2, 0.1388)	(0.1, 0.016)	};
\addplot+[] coordinates {	(1, 0.5744)	(0.5, 0.5758)	(0.4, 0.5788)	(0.3, 0.5162)	(0.2, 0.3496)	(0.1, 0.1045)	};
\addplot+[] coordinates {	(1, 0.5744)	(0.5, 0.5038)	(0.4, 0.3913)	(0.3, 0.2706)	(0.2, 0.1646)	(0.1, 0.042)	};
\addplot+[] coordinates {	(1, 0.5744)	(0.5, 0.4852)	(0.4, 0.4423)	(0.3, 0.4009)	(0.2, 0.323)	(0.1, 0.1458)	};
\addplot+[] coordinates {	(1, 0.5744)	(0.5, 0.4729)	(0.4, 0.4184)	(0.3, 0.2771)	(0.2, 0.1111)	(0.1, 0.0126)	};
							
\nextgroupplot[title={Paris6k},
    ymax=0.8,]							
\addplot+[] coordinates {	(1, 0.7235)	(0.5, 0.6972)	(0.4, 0.6682)	(0.3, 0.5887)	(0.2, 0.4701)	(0.1, 0.3408)	};
\addplot+[] coordinates {	(1, 0.7235)	(0.5, 0.5549)	(0.4, 0.4684)	(0.3, 0.4126)	(0.2, 0.3734)	(0.1, 0.1626)	};
\addplot+[] coordinates {	(1, 0.7235)	(0.5, 0.6412)	(0.4, 0.5535)	(0.3, 0.4934)	(0.2, 0.4379)	(0.1, 0.2592)	};
\addplot+[] coordinates {	(1, 0.7235)	(0.5, 0.5154)	(0.4, 0.4755)	(0.3, 0.3329)	(0.2, 0.2241)	(0.1, 0.0404)	};
\addplot+[] coordinates {	(1, 0.7602)	(0.5, 0.7403)	(0.4, 0.723)	(0.3, 0.6793)	(0.2, 0.5522)	(0.1, 0.29)	};
\addplot+[] coordinates {	(1, 0.7602)	(0.5, 0.6627)	(0.4, 0.5531)	(0.3, 0.4199)	(0.2, 0.3042)	(0.1, 0.0915)	};
\addplot+[] coordinates {	(1, 0.7602)	(0.5, 0.7015)	(0.4, 0.6476)	(0.3, 0.5883)	(0.2, 0.5202)	(0.1, 0.1919)	};
\addplot+[] coordinates {	(1, 0.7602)	(0.5, 0.6268)	(0.4, 0.5655)	(0.3, 0.3709)	(0.2, 0.1751)	(0.1, 0.0359)	};
							
\nextgroupplot[
    title={UKbench},
    y label style={at={(axis description cs:0.09,0.8)},anchor=east},
    ylabel={$4\times$recall@4},
    yticklabel={\pgfmathprintnumber{\tick}},
    ymin=0, ymax=4
]
\addplot+[] coordinates {	(1, 3.7561)	(0.5, 3.7458)	(0.4, 3.7098)	(0.3, 3.5827)	(0.2, 3.2365)	(0.1, 2.8954)	};
\addlegendentry{H. 1, SQP}

\addplot+[] coordinates {	(1, 3.7561)	(0.5, 3.4196)	(0.4, 3.2767)	(0.3, 3.0034)	(0.2, 2.644)	(0.1, 1.9709)	};
\addlegendentry{H. 2, SQP}

\addplot+[] coordinates {	(1, 3.7561)	(0.5, 3.6359)	(0.4, 3.554)	(0.3, 3.4325)	(0.2, 3.1769)	(0.1, 2.5727)	};
\addlegendentry{H. 3, SQP}

\addplot+[] coordinates {	(1, 3.7561)	(0.5, 3.4811)	(0.4, 3.3522)	(0.3, 3.0336)	(0.2, 2.3224)	(0.1, 0.7404)	};
\addlegendentry{H. 4, SQP}

\addplot+[] coordinates {	(1, 3.7446)	(0.5, 3.7322)	(0.4, 3.6975)	(0.3, 3.6002)	(0.2, 3.3345)	(0.1, 2.7583)	};
\addlegendentry{H. 1, R-MAC}

\addplot+[] coordinates {	(1, 3.7446)	(0.5, 3.4887)	(0.4, 3.3192)	(0.3, 2.9287)	(0.2, 2.249)	(0.1, 1.4023)	};
\addlegendentry{H. 2, R-MAC}

\addplot+[] coordinates {	(1, 3.7446)	(0.5, 3.6125)	(0.4, 3.542)	(0.3, 3.4345)	(0.2, 3.1503)	(0.1, 2.156)	};
\addlegendentry{H. 3, R-MAC}

\addplot+[] coordinates {	(1, 3.7446)	(0.5, 3.4568)	(0.4, 3.3119)	(0.3, 2.7686)	(0.2, 1.8549)	(0.1, 0.9303)	};
\addlegendentry{H. 4, R-MAC}

\end{groupplot}
}\end{tikzpicture}

\end{minipage}
\caption{Performance of pruned networks by remaining network size across different datasets. In each dataset, lines are colored by pruning heuristic and dashed by pooling feature (solid lines use SQP pooling, dashed lines use R-MAC). No fine-tuning was performed. We observe that Heuristic 1 consistently performs better than the other heuristics, and the use of SQP features over R-MAC improves performance.}
\label{fig:res_overview_noft}
\end{figure*}

%% file: results_finetuning.tex
\begin{figure}[htb]
\begin{minipage}[b]{1.0\linewidth}
  \centering
  \begin{tikzpicture}{
  \pgfplotsset{every x tick label/.append style={font=\tiny}}
  \pgfplotsset{every y tick label/.append style={font=\small}}

\pgfplotscreateplotcyclelist{custom_bar_color_list}{%
fill=orange,fill opacity=0.25,text opacity=1\\%
fill=black,fill opacity=0.0,text opacity=1,thick\\%
}

\begin{groupplot}[
    group style={
        group name=my plots,
        group size=4 by 1,
        xlabels at=edge bottom,
        xticklabels at=edge bottom,
        vertical sep=0pt
        ylabels at=edge left,
        yticklabels at=edge left,
        horizontal sep=0pt
    },
    title={Title},
    width = 1.68cm, height = 3cm,
    scale only axis,
    symbolic x coords={H1, H2, H3, H4},
    x tick label style = {font=\tiny,anchor=north},
    y tick label style = {font=\tiny,anchor=east},
    xtick = data,
    enlarge x limits=0.18,
    ybar stacked,
    /pgf/bar width=9pt,
    ymajorgrids,
    point meta={y*100},
    every node near coord/.append style={font=\scriptsize},
    yticklabel={\pgfmathparse{\tick*100}\pgfmathprintnumber{\pgfmathresult}\%},
    ytick pos=left,
    ymin=0.4, ymax=1,
    cycle list name=custom_bar_color_list,
    legend cell align=left,
    legend columns=5,
    legend style={
        at={(0.5,-0.3)},
        anchor=north,
        column sep=1ex
    }
  ]

\nextgroupplot[title={Holidays},
  y label style={at={(axis description cs:0.35,0.5)},anchor=south},
  ylabel={MAP}
]
\addplot+[
  point meta=explicit symbolic,
  nodes near coords,
] coordinates {
(H1, 0.8642)[-2]
(H2, 0.7406)
(H3, 0.8265)[-2]
(H4, 0.7907)
};
\addplot+[
  point meta=explicit symbolic,
  nodes near coords,
] coordinates {
(H1, -0.0238)
(H2, 0.0536)[+5]
(H3, -0.021)
(H4, 0.0017)[0]
};

\nextgroupplot[title={Oxford5k}]
\addplot+[] coordinates {
(H1, 0.6063)
(H2, 0.4179)
(H3, 0.5164)
(H4, 0.4772)
};
\addplot+[
  point meta=explicit symbolic,
  nodes near coords,
] coordinates {
(H1, 0.1646)[+16]
(H2, 0.2688)[+27]
(H3, 0.2178)[+22]
(H4, 0.2121)[+21]
};

\nextgroupplot[title={Paris6k}]
\addplot+[] coordinates {
(H1, 0.6972)
(H2, 0.5549)
(H3, 0.6412)
(H4, 0.5154)
};
\addplot+[
  point meta=explicit symbolic,
  nodes near coords,
] coordinates {
(H1, 0.049)[+5]
(H2, 0.1741)[+17]
(H3, 0.1006)[+10]
(H4, 0.2026)[+20]
};

\nextgroupplot[title={UKbench},
  axis y line*=right,
  yticklabel pos=right,
  ytick={1.6, 2.4, 3.2, 4},
  yticklabels={1.6, 2.4, 3.2, 4.0},
  clip=false,
  y tick label style={at={(axis description cs:0.5,0.5)},anchor=west},
  ylabel near ticks,
  y label style={at={(axis description cs:1.2,0.5)},anchor=north},
  ylabel={$4\times$recall@4},
  ymin=1.6,
  ymax=4,
]
\addplot+[
  point meta=explicit symbolic,
  nodes near coords,
] coordinates {
(H1, 3.7458)[-.33]
(H2, 3.4196)[-.11]
(H3, 3.6359)[-.34]
(H4, 3.4811)[-.25]
};
\addplot+[] coordinates {
(H1, -0.3287)
(H2, -0.1113)
(H3, -0.3375)
(H4, -0.2477)
};

\end{groupplot}
}\end{tikzpicture}

\end{minipage}
\caption{Performance of 50\% pruned networks on different heuristics, grouped by dataset. The colored bar represents performance before fine-tuning, and the clear bar represents the change in performance with fine-tuning for 20 epochs. The number above each bar is the change in percentage points or value. We observe that fine-tuning greatly improves the performance across \texttt{Oxford5k} and \texttt{Paris6k} datasets, suggesting that it is an important step in our pruning pipeline. }
\label{fig:finetuning}
\end{figure}
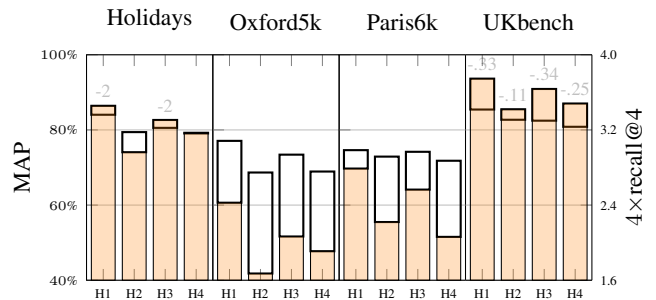

%% file: results_sh_threshold.tex
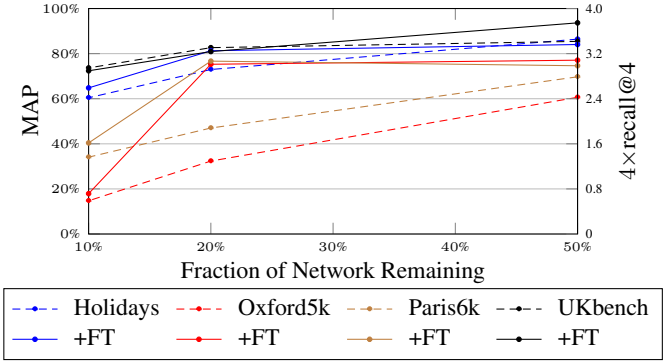
\begin{figure}[tb]
\begin{minipage}[b]{1.0\linewidth}
  \centering
  \begin{tikzpicture}{

\pgfplotscreateplotcyclelist{custom_color_list}{
blue,densely dashed,mark=*,mark size=0.85\\%
red,densely dashed,mark=*,mark size=0.85\\%
brown,densely dashed,mark=*,mark size=0.85\\%
black,densely dashed,mark=*,mark size=0.85\\%
blue,mark=*,mark size=0.85\\%
red,mark=*,mark size=0.85\\%
brown,mark=*,mark size=0.85\\%
black,mark=*,mark size=0.85\\%
}

\begin{axis}[
    width = 6.5cm, height = 3cm,
    scale only axis,
    xticklabel={\pgfmathparse{\tick*100}\pgfmathprintnumber{\pgfmathresult}\%},
    x tick label style = {font=\tiny,anchor=north},
    /pgf/bar width=7pt,
    ymajorgrids,
    x label style={at={(axis description cs:0.5,0.09)},anchor=north,font=\small},
    xlabel={Fraction of Network Remaining},
    y label style={at={(axis description cs:0.07,0.7)},anchor=east,font=\small},
    ylabel={MAP},
    point meta={y*100},
    y tick label style={font=\tiny},
    yticklabel={\pgfmathparse{\tick*100}\pgfmathprintnumber{\pgfmathresult}\%},
    ytick pos=left,
    xmin=0.1, xmax=0.5,
    ymin=0, ymax=1,
    cycle list name=custom_color_list,
    legend cell align=left,
    legend columns=4,
    legend style={
        anchor=north,
        at={(0.5,-0.24)},
        column sep=1ex
    }
  ]
	
\addplot+[] coordinates {	( 0.5,0.8642)	( 0.2,0.7296)	( 0.1,0.6042)	};
\addlegendentry{Holidays}
\addplot+[] coordinates {	( 0.5,0.6063)	( 0.2,0.3235)	( 0.1,0.1473)	};
\addlegendentry{Oxford5k}
\addplot+[] coordinates {	( 0.5,0.6972)	( 0.2,0.4701)	( 0.1,0.3408)	};
\addlegendentry{Paris6k}
\addplot+[] coordinates {	( 0.5,0.854275)	( 0.2,0.8265)	( 0.1,0.7365)	};
\addlegendentry{UKbench}

\addplot+[] coordinates {	( 0.5,0.8404)	( 0.2,0.8127)	( 0.1,0.648)	};
\addlegendentry{+FT}
\addplot+[] coordinates {	( 0.5,0.7709)	( 0.2,0.7528)	( 0.1,0.179)	};\addlegendentry{+FT}
\addplot+[] coordinates {	( 0.5,0.7462)	( 0.2,0.7667)	( 0.1,0.404)	};
\addlegendentry{+FT}
\addplot+[] coordinates {	( 0.5,0.93645)	( 0.2,0.809125)	( 0.1,0.72385)	};
\addlegendentry{+FT}

\end{axis}

\begin{axis}[
  width = 6.5cm, height = 3cm,
  scale only axis,
  axis y line*=right,
  axis x line=none,
  ymin=0, ymax=4,
  ytick={0, 0.8, 1.6, 2.4, 3.2, 4.0},
  yticklabels={0, 0, 0.8, 1.6, 2.4, 3.2, 4.0},
  y tick label style={font=\tiny},
  ylabel={$4\times$recall@4},
  ylabel near ticks,
]

\end{axis}

}\end{tikzpicture}

\end{minipage}
\caption{Performance (with SQP pooling) of networks pruned by heuristic 1, before and after fine-tuning (FT). Pruning the network reduces performance significantly, but fine-tuning can generally restore most of the performance.}
\label{fig:res_sh_threshold}
\end{figure}

%% file: layersize.tex
\begin{figure}[tb]
\begin{minipage}[b]{1.0\linewidth}
\centering
\begin{tikzpicture}{

\pgfplotscreateplotcyclelist{custom_orange_color_list}{%
  draw=black,mark=*,mark size=0.85\\%
  draw=red,mark=*,mark size=0.85\\%
  draw=brown,mark=*,mark size=0.85\\%
}

\begin{axis}[
  scale only axis,
  width = 7cm, height = 4cm,
  symbolic x coords={c11,	c12,	c21,	c22,	c31,	c32,	c33,	c41,	c42,	c43,	c51,	c52,	c53},
  x label style={at={(axis description cs:0.5,0.05)},anchor=north},
  x tick label style = {font=\tiny,anchor=north},
  xlabel={Network Layer},
  ytick pos=left,
  ymajorgrids=true, ymin=0, ymax=1,
  y label style={at={(axis description cs:0.05,0.5)},anchor=north},
  ylabel={Fraction layer remaining},
  y tick label style = {font=\tiny,anchor=east},
  yticklabel={\pgfmathparse{\tick*100}\pgfmathprintnumber{\pgfmathresult}\%},
  % x tick label style = {rotate=35,anchor=west,xshift=-5pt,yshift=4pt},
  /pgf/bar width=14pt,
  enlarge x limits=0.05,
  legend style={font=\tiny},
  cycle list name=custom_orange_color_list,
]

\addplot+[] coordinates {	(c11, 0.974)	(c12, 0.827)	(c21, 0.763)	(c22, 0.704)	(c31, 0.655)	(c32, 0.586)	(c33, 0.594)	(c41, 0.522)	(c42, 0.436)	(c43, 0.45)	(c51, 0.496)	(c52, 0.519)	(c53, 0.497)	};
\addlegendentry{50\% Network Remaining}

\addplot+[] coordinates {	(c11, 0.96)	(c12, 0.678)	(c21, 0.578)	(c22, 0.489)	(c31, 0.405)	(c32, 0.304)	(c33, 0.31)	(c41, 0.224)	(c42, 0.139)	(c43, 0.15)	(c51, 0.189)	(c52, 0.205)	(c53, 0.188)	};
\addlegendentry{20\% Network Remaining}

\addplot+[] coordinates {	(c11, 0.95)	(c12, 0.59)	(c21, 0.472)	(c22, 0.378)	(c31, 0.286)	(c32, 0.183)	(c33, 0.188)	(c41, 0.118)	(c42, 0.057)	(c43, 0.065)	(c51, 0.086)	(c52, 0.094)	(c53, 0.085)	};
\addlegendentry{10\% Network Remaining}

\end{axis}

}\end{tikzpicture}

\end{minipage}
\caption{Fraction of each layer remaining after different amounts of pruning (with heuristic 1). We observe that the layers closest to the data are pruned the least, a pattern that is consistent with all heuristics. This is likely because our pruning leverages the increasing redundancy in higher layers.}
\label{fig:layersize}
\end{figure}
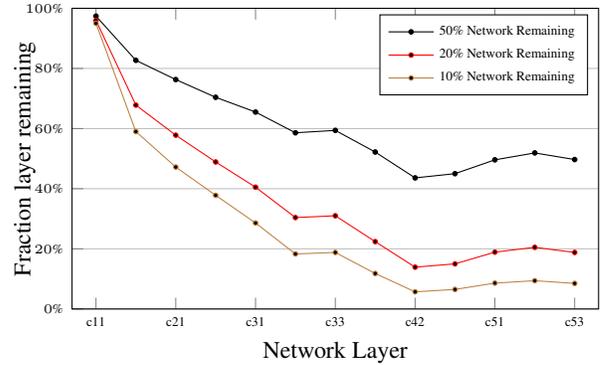